\documentclass[10pt,twocolumn,letterpaper]{article}
\usepackage[pagenumbers]{cvpr}
\usepackage{algorithm}
\usepackage{algorithmic}
\usepackage{pifont}
\usepackage{multirow}
\usepackage{tcolorbox}

\definecolor{lightroyalblue}{HTML}{F6F8FD}
\newtcolorbox{mybox}{
  colback=lightroyalblue,
  colframe=black,
  boxsep=0pt,
  left=4pt,right=4pt,top=4pt,bottom=4pt
}

\makeatletter
\let\cvpr@maketitle@original\@maketitle
\def\@maketitle{%
  \cvpr@maketitle@original
  \begin{center}
    \includegraphics[width=\textwidth]{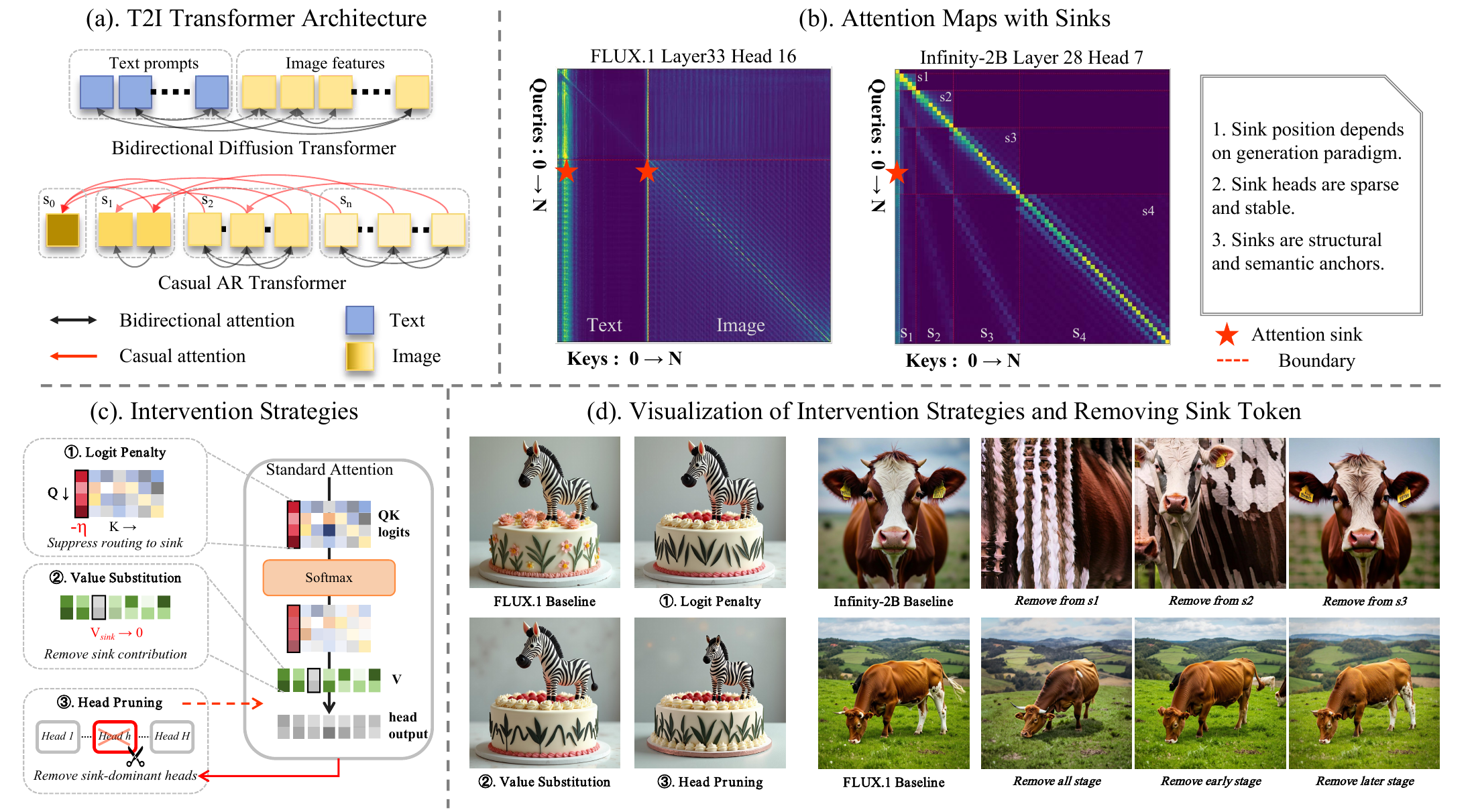}
    \captionof{figure}{Overview of our analysis on attention sinks in text-to-image Transformers.
    (a) Transformer architecture of bidirectional diffusion and causal autoregressive models;
    (b) representative sink patterns in FLUX.1 (Diffusion) and Infinity-2B (Autoregressive);
    (c) intervention strategies, including logit penalty, value substitution, and head pruning;
    (d) their effects together with stage-wise global-token removal.
    The evaluated diffusion models concentrate attention near modality boundaries and spatial corner positions, whereas Infinity-2B exhibits strong raw concentration at its first/global token.}
    \label{fig:showcase}
  \end{center}
}
\makeatother

\definecolor{cvprblue}{rgb}{0.21,0.49,0.74}
\usepackage[pagebackref,breaklinks,colorlinks,allcolors=cvprblue]{hyperref}

\title{Exploring Attention Sinks in Text-to-Image Transformers from Diffusion and Autoregressive Models}

\author{Binglei Li\textsuperscript{1,2},~ Mengping Yang\textsuperscript{1,3~$\dagger$},~ Zhiyu Tan\textsuperscript{1,3},
~ Wenxu Qian\textsuperscript{1,2}, ~ Junping Zhang\textsuperscript{1~$\#$}, ~ Hao Li\textsuperscript{1,2,3~$\#$}\\
\textsuperscript{1}Fudan University \quad \textsuperscript{2}
Shanghai Innovation Institute \quad \textsuperscript{3} 
Shanghai Academy of AI for Science\\
\vspace{3pt}
{ \textsuperscript{$\dagger$}Project Lead \qquad
\textsuperscript{$\#$}Corresponding Authors}\\
\vspace{5pt}
\texttt{\{blli24\}@m.fudan.edu.cn, \{jpzhang,lihao\_lh\}@fudan.edu.cn}
}

\begin{document}
\maketitle

% !TEX root = ../main.tex

\begin{abstract}
%%%%% from mengping %%%%%
%
Attention sinks are tokens or positions that consistently receive disproportionate attention mass, often beyond what their semantic content would suggest. 
Despite extensive study in large language models, their behavior and functional relevance in text-to-image (T2I) Transformers remain largely unexplored.
To fill this gap and better understand how text-to-image Transformers allocate and route attention, we explore this phenomenon of text-to-image transformers across bidirectional diffusion models and causal autoregressive models.
Specifically, we explore \emph{whether attention sinks exist}, \emph{where they emerge}, and \emph{how they functionally influence generation}.
Through systematic attention analysis and targeted interventions, we identify that attention sinks emerge in both bidirectional diffusion and autoregressive models but exhibit distinct patterns.
The autoregressive model concentrates attention on the first and global-scale tokens, whereas the bidirectional models develop sinks near modality boundaries, within text regions, and at spatial corners.
Moreover, these patterns vary across generation stages and are localized to a sparse subset of layers and heads.
To assess their functional relevance, we perturb sink-related heads and token positions.
It turns out that perturbing sink-ranked heads produces larger average changes than non-sink comparisons, while some sink-position perturbations have limited effects on bidirectional model scores.
We further introduce register tokens to redistribute attention sinks and find that several configurations preserve or modestly improve generation performance.
Overall, attention sinks are architecture-, inference-stage-, and component-dependent phenomena, providing a basis for mechanistic analysis and principled attention design in text-to-image transformers.

\end{abstract}

% !TEX root = ../main.tex

\section{Introduction}
\label{sec:intro}

Transformers have become a central architecture for modern generative modeling, but their attention maps often contain \emph{attention sinks}, where a small set of token positions absorb disproportionately large attention mass~\cite{attention_vaswani2017,xiao2024efficient}.
Such phenomenon has been extensively studied in large language models (LLMs), where causal masking and sequence organization often cause early tokens to act as persistent attention anchors~\cite{xiao2024efficient,sun2024massive,gu2025empirialview,guo2024attention}.
However, the prevailing understanding of attention sinks remains largely centered in LLMs, with limited exploration in other domains.
%
% Related sink-like patterns have also been reported in Vision Transformers and multimodal LLMs, where high-attention tokens may correspond to low-information image patches or visual positions that affect grounding behavior~\cite{bondarenko2023quantizable,darcet2024vision,kang2025see,zhang2025shallow,luo2025sink}.
%

Text-to-image Transformers differ substantially from LLMs.
They jointly process text and image tokens, organize visual tokens according to spatial or multiscale structures, and generate images either through iterative diffusion denoising or autoregressive image-token prediction~\cite{SD3_esser2024scaling,flux_2024,cai2025zimage,tian2024var,han2025infinity}.
Therefore, their attention distributions might be relevant to modality boundaries, spatial topology, attention directionality, and generation stage.
These properties may not be reflected by the conventional first-token sink of causal LLMs.
Moreover, sink patterns reported in vision models~\cite{zhang2025shallow,luo2025sink} do not directly explain how attention is allocated during multimodal T2I generation.
A systematic study of attention sinks in T2I models is thus needed not only to determine whether the phenomenon occurs beyond LLMs, but also to better understand how T2I Transformers allocate and route attention during generation.

This paper addresses three fundamental questions: \emph{1) whether attention sinks exist in modern T2I Transformers, 2) where they are located, and 3) are they functionally relevant to generation?}
To answer them, we conduct a systematic empirical study across three models: FLUX.1 and Z-Image as diffusion-based T2I Transformers, and Infinity-2B as an autoregressive T2I Transformer, as shown in Fig.~\ref{fig:showcase}(a).
We first characterize their attention distributions by sink strength, attention inequality, layer-head localization, positional preference, prompt sensitivity, and inference-time dynamics.
Concretely, we evaluate these properties on GenEval~\cite{geneval_ghosh2023} and DPG-Bench~\cite{dpg_hu2024ella}, which cover diverse compositional generation and fine-grained text-image alignment.
Then, we perform targeted interventions on attention routing, value contributions, sink-related heads, and candidate sink positions to assess their functional relevance.
Finally, we introduce register tokens as dedicated non-semantic positions to examine whether sink-related attention can be redistributed without comprising synthesis quality.

Our results show that attention-sink patterns are structured in both diffusion and autoregressive T2I transformers, yet their locations differ substantially across architectures.
Autoregressive model exhibits pronounced concentration on the global first token, while the diffusion models concentrate attention near modality boundaries, within text-token regions, and at spatial corners.
Moreover, sink behavior is localized to a sparse subset of layers and heads and evolves across denoising or coarse-to-fine generation stages, suggesting that sink locations of T2I transformers are associated with the inference stage rather than fixed token positions.

To investigate their influence on generation, we perform interventions on sink tokens, which reveal heterogeneous functional effects.
Perturbing sink-ranked heads produces larger average changes than the reported random or low-sink comparisons, indicating that certain sink-related heads participate in consequential generation pathways.
In comparison, several sink-position perturbations have limited effects in bidirectional diffusion models.
Additionally, register-token experiments further show that several configurations can redistribute attention while preserving or modestly improving generation performance.
Together, these results extend the study of attention sinks beyond its language-centric formulation.
Our analysis characterizes attention sinks as architecture-, stage-, and component-dependent phenomena and establishes an important distinction between where attention accumulates and where generation is functionally sensitive.
Our contributions are summarized as:
\begin{itemize}
    \item We systematically explore and characterize attention sinks in bidirectional diffusion and causal autoregressive T2I Transformers across token positions, layers, heads, prompts, and generation stages.
    \item We evaluate the functional relevance of sink-related components through targeted interventions and investigate register tokens as dedicated non-semantic positions for redistributing attention.
    \item We provide an empirical foundation for controlled mechanistic analysis and for the principled design of attention mechanisms and token layouts in T2I Transformers.
\end{itemize}

% !TEX root = ../main.tex

\section{Preliminaries}
\label{sec:preliminaries}

\noindent \textbf{Attention Sinks}.
An \emph{attention sink} (AS) is a token position $j$ that consistently receives disproportionately high cumulative attention across queries and heads~\cite{xiao2024efficient,sun2024massive}.
For each attention head $h \in H$ at layer $l\in L$ and timestep $t$, the cumulative \emph{attention mass} of token $j$ is:
\begin{equation}
    \hat{\mathbf{A}}^{(t,l,h)}_{j} = \sum\nolimits_{i=1}^{N} \mathbf{A}^{(t,l,h)}_{i,j} \big/ N , \quad \mathbf{A}^{(t,l,h)} \in \mathbb{R}^{N \times N}.
    \label{eq:attention_mass}
\end{equation}
The set of AS tokens is then given by:
\begin{equation}
    \mathcal{S}_{AS}=\{j^*| \hat{\mathbf{A}}_{j^*} > \tau \cdot \mu_{\mathbf{A}}\},~ \mu_{\mathbf{A}}=\tfrac{1}{N}\sum\nolimits_{k=1}^{N}\hat{\mathbf{A}}_{k},
\end{equation}
where $\tau$ is a concentration threshold.
This formulation highlights the extreme numerical prominence of AS tokens.
Across different architectures, AS tokens consistently correspond to uninformative tokens.

\noindent \textbf{Text-to-Image Transformers}.
The Transformer~\cite{attention_vaswani2017} defines a sequence modeling paradigm based on the multi-head self-attention mechanism.
For an input $\mathbf{X}\in \mathbb{R}^{N\times D}$, where $N$ denotes the sequence length and $D$ the feature dimension, the queries $\mathbf{Q}$, keys $\mathbf{K}$, and values $\mathbf{V}$ are obtained via linear projection.
Attention is then computed:
\begin{equation}
    \text{Attention}(\mathbf{Q}, \mathbf{K}, \mathbf{V})=\text{Softmax}(\tfrac{\mathbf{QK}^T}{\sqrt{d_k}})\mathbf{V}.
    \label{eq:attention}
\end{equation}
where $d_k = D/H$ and $H$ is the total number of heads.
In \emph{MMDiTs}~\cite{SD3_esser2024scaling,flux_2024,cai2025zimage}, the image feature $\mathbf{X}_\text{img}$ and the text embedding $\mathbf{X}_\text{txt}$ are concatenated and jointly modeled via bidirectional attention across $L$ transformer blocks to progressively denoise the latent.
\emph{Autoregressive models}~\cite{han2025infinity,liu2026infinitystar,tian2024var} factorize the joint distribution as $p(\mathbf{X})=\prod_i p(\mathbf{X}_i\mid \mathbf{X}_{<i})$, with the transformers predicting the next token via causal attention or next scale over a coarse-to-fine hierarchy.

% !TEX root = ../main.tex

% \section{Characterizing Attention Sinks in T2I Transformers}
\section{Analytical Framework}
\label{sec:characterization}
We establish a unified framework for identifying, localizing, and intervening on attention sinks in T2I Transformers, with related metrics (Sec.~\ref{sec:sink_metrics}), intervention strategies (Sec.~\ref{sec:intervention_strategy}), and experimental setup (Sec.~\ref{sec:setup}).

\subsection{Metrics for Attention Sinks}
\label{sec:sink_metrics}

\noindent \textbf{Sink Concentration}.
Unless otherwise specified, we average cumulative attention mass over prompts and generation steps.
For head $h$ at layer $l$, we define its \emph{sink concentration} as
\begin{equation}
    \kappa^{(l,h)}
    = \bigl( \max_{j}\;\hat{\mathbf{A}}^{(l,h)}_{j} \bigr) \big/ \bigl( \sum\nolimits_{j}\hat{\mathbf{A}}^{(l,h)}_{j} \bigr),
    \label{eq:sink_concentration}
\end{equation}
where $\hat{\mathbf{A}}^{(l,h)}_{j}$ denotes the averaged attention mass received by token $j$, defined in Eq.~\ref{eq:attention_mass}.
Because $\hat{\mathbf{A}}^{(l,h)}_{j}$ is normalized by the number of visible tokens, the uniform-attention null of $\kappa$ is mask-independent, giving $\kappa^{(l,h)}\!\approx\!1/N$ under bidirectional, causal, and variable-availability masks alike.
Values above $1/N$ therefore indicate genuine concentration on a few tokens rather than mask geometry, and $\kappa$ is directly comparable across the two models.
We rank heads by $\kappa^{(l,h)}$ and denote the top-$K$ sink heads as $\mathcal{H}_{\mathrm{sink}}$.
For each head, the top-$k$ sink-token set is 
\begin{equation}
    \mathcal{S}^{(l,h)}_{k}=\left\{j^* \mid \mathrm{Top}_k\bigl(\hat{\mathbf{A}}^{(l,h)}_{j}\bigr) > \tau \cdot \mu_{\mathbf{A}} \right\},
    \label{eq:sink_token_set}
\end{equation}
which contains the $k$ token positions receiving the largest attention mass.
Empirically, we set $\tau=10$ for the diffusion models and $\tau=100$ for Infinity-2B.

\noindent \textbf{Position-Wise Sink Ratio}.
For each evaluated attention unit $u\in\mathcal{U}$, let $\mathcal{S}^{(u)}_k$ denote its top-$k$ sink-token set.
For a candidate position $c$, such as the sequence start, a modality boundary, or an autoregressive global-scale token, we define its \emph{position-wise sink ratio} as
\begin{equation}
    \rho_k(c)
    =
    \frac{1}{|\mathcal{U}|}
    \sum_{u\in\mathcal{U}}
    \mathbf{1}\!\left[c\in\mathcal{S}^{(u)}_k\right],
    \label{eq:position_sink_ratio}
\end{equation}
where $\mathcal{U}$ is the set of layer--head units included in the analysis.
Thus, $\rho_1(c)$ is the fraction of units for which $c$ is the dominant sink, whereas $\rho_5(c)$ measures how often $c$ appears among the five strongest sinks.
This metric compares the prominence of predefined positions, but by itself does not distinguish a sink-specific effect from causal visibility or an architectural token role.

\noindent \textbf{Generation-Time Dynamics}.
We divide inference into early, middle, and late stages.
For diffusion models, $R_{\mathrm{img}}$ measures the fraction of top-$k$ sink tokens falling in the image-token region.
For Infinity-2B (autoregressive coarse-to-fine), we use a fine-scale score $R_{\mathrm{fine}}$ to measure whether sink mass shifts toward finer visual scales.
We quantify positional stability via the Jaccard overlap (i.e., the ratio of the intersection to the union) of the top‑$k$ sink sets across adjacent steps and across generation stages.
Detailed formulas for $R_{\mathrm{img}}$, $R_{\mathrm{fine}}$ are provided in the appendix.

\noindent \textbf{Prompt-Distribution Stability}.
To assess whether sink behaviors depend on prompt distribution, we conduct experiments on GenEval and DPG-Bench.
We compare the sink-head stability using Pearson correlation of two sets of sink-head profiles.
To measure whether the sink position of a fixed attention unit wanders across prompts, we use normalized sink-position entropy $H_{\mathrm{norm}}$ and report the fraction of high-entropy units with $H_{\mathrm{norm}}\ge 0.5$.
Detailed formulas and explanations are provided in the appendix.

\subsection{Intervention Strategy}
\label{sec:intervention_strategy}

To probe the functional role of attention sinks, we use three suppressive interventions targeting different components of multi-head attention:
\begin{equation}
    \mathbf{x} \!\leftarrow\! \mathbf{x}\! +\! \sum_{h=1}^{H} m^{(h)}\! \cdot\!
    \mathrm{Softmax}\!\left(\frac{\mathbf{Q}^{(h)}\mathbf{K}^{(h)\top}}{\sqrt{d_k}}\!+\! \mathbf{B}^{(h)}\right) \tilde{\mathbf{V}}^{(h)},
    \label{eq:unified_intervention}
\end{equation}
where $m^{(h)}\!\in\!\{0,1\}$ is a head mask, $\mathbf{B}^{(h)}\!\in\!\mathbb{R}^{N\times N}$ is a logit-bias matrix, and $\tilde{\mathbf{V}}^{(h)}$ is a possibly modified value matrix.
The vanilla model sets $m^{(h)}\!=\!1$, $\mathbf{B}^{(h)}\!=\!\mathbf{0}$, and $\tilde{\mathbf{V}}^{(h)}\!=\!\mathbf{V}^{(h)}$ for all heads.
During interventions, we apply only one intervention strategy at a time. 

\noindent \textbf{Logit Penalty}.
Given sink positions $\mathcal{S}^{(l,h)}_{k}$, we set $B^{(h)}_{ij}=-\eta\,\mathbf{1}[j\in\mathcal{S}^{(l,h)}_{k}]$ with penalty strength $\eta>0$. Typically, we set $\eta=10$.
This perturbs attention routing before the softmax while leaving value states unchanged.

\noindent \textbf{Value Substitution}.
We zero out value vectors at sink positions, i.e., $\tilde{\mathbf{V}}_{j}=\mathbf{0}$ for $j\in\mathcal{S}^{(l,h)}_{k}$ and $\tilde{\mathbf{V}}_{j}=\mathbf{V}_{j}$ otherwise.
This tests whether sink-token value states carry useful information even when attention routing remains intact.

\noindent \textbf{Head Pruning}.
We set $m^{(h)}=0$ for $h\in\mathcal{H}_{\mathrm{sink}}$, removing the entire output contribution of selected sink heads.
This measures how generation changes when the output contribution of selected high-concentration heads is removed.

\subsection{Experimental Setups}
\label{sec:setup}

\noindent \textbf{Models}.
We study two T2I Transformer paradigms: bidirectional denoising models, including FLUX.1-dev~\cite{flux_2024} and Z-Image~\cite{cai2025zimage}, and a causal autoregressive model, Infinity-2B~\cite{han2025infinity}.
During inference, all models generate $1024\!\times\!1024$ images using their official inference settings.
FLUX.1 and Z-Image use the EDM~\cite{karras2022elucidating} sampler with $28$ denoising steps, while Infinity-2B uses a bitwise sampler with a $13$-scale schedule.

\noindent \textbf{Token Layouts}.
Token layouts differ across models: Infinity follows a causal sequence, whereas FLUX.1 and Z-Image adopt a fixed interleaved structure.
Moreover, FLUX.1 places text tokens first and Z-Image places image tokens first. 
We therefore define three positional relations: 1) sequence position, 2) text-side relation, and 3) image-side relation.
In the following, \emph{token position} refers to the index within the sequence, while modality relations are discussed separately.
%
%
%
% Because token layouts differ across models, we interpret sink positions using three positional relations: 1) sequence position, 2) text-side relation, and 3) image-side relation.
% %
% Here, \emph{token position} refers to the sequence index, while modality relations are analyzed separately.

\noindent \textbf{Datasets and Metrics}.
We evaluate on two complementary benchmarks: GenEval~\cite{geneval_ghosh2023}, which contains $553$ prompts for compositional generation such as counting, color attribution, and spatial relations, and DPG-Bench~\cite{dpg_hu2024ella}, which contains $1065$ prompts for fine-grained text-image alignment.
Together, they provide $1618$ prompts with distinct lengths and semantic densities.
We report official GenEval and DPG-Bench scores for semantic alignment, and use HPSv2~\cite{HPSV2_wu2023human} and ImageReward~\cite{imagereward_xu2023} as auxiliary preference and perceptual-quality metrics.
For intervention tables, $\bar{\Delta}$ denotes the average percentage change across the reported metrics relative to the corresponding baseline.
All evaluations follow the official protocols, and all experiments are conducted on NVIDIA H100 GPUs.

% !TEX root = ../main.tex

\section{Systematic Analysis}
\label{sec:systematic_analysis}

\begin{table}[!t]
    \centering
    % \normalsize
    \small
    \caption{Sink ratios of predefined candidate positions.
    Infinity exhibits pronounced concentration on the first token, while the diffusion models concentrate attention near modality boundaries, in text-token regions, and at spatial corners.
    }
    \setlength{\tabcolsep}{4.0mm}
    \begin{tabular}{@{}llcc@{}}
        \toprule
        Model & Candidate $c$ & $\rho_1(c)$ & $\rho_5(c)$ \\
        \midrule
        \multirow{3}{*}{FLUX.1} & sequence pos 0 & 0.32\% & 0.56\% \\
         & text pos 7 & 12.95\% & 32.26\% \\
         & boundary pos 511 & 6.05\% & 23.06\% \\
         & image pos 4544 & 3.04\% & 18.33\% \\
        \cline{2-4}
        \multirow{3}{*}{Z-Image} & sequence pos 0 & 0.47\% & 0.65\% \\
         & boundary pos 4096 & 10.89\% & 21.31\% \\
         & text pos 4109 & 11.62\% & 32.49\% \\
        \cline{2-4}
        Infinity-2B & 1$\times$1 token pos 0 & 37.52\% & 100.00\% \\
        \bottomrule
    \end{tabular}
    
    \label{tab:first_token_ratio}
\end{table}

\begin{figure}[!t]
    \centering
    \includegraphics[width=\linewidth]{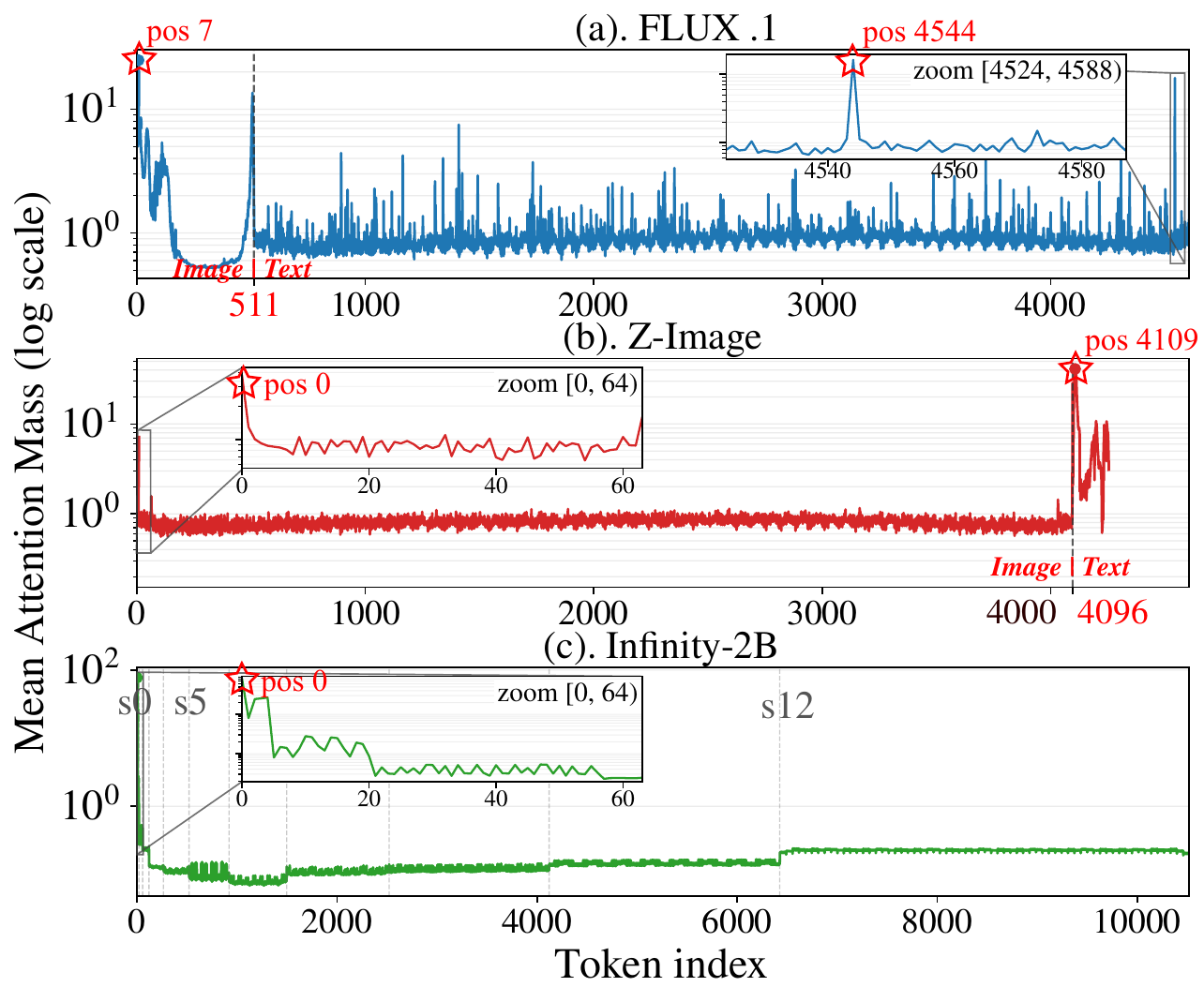}
    \caption{Attention mass by token position across models.
    We highlight the most prominent sink positions for each model.
    Panels show FLUX.1, Z-Image, and Infinity-2B, while vertical markers identify modality or scale boundaries and insets magnify local peaks.
    Diffusion models peak near modality boundaries or image corners, whereas Infinity exhibits a pronounced position-0 sink.
    }
    \label{fig:attention_mass_distribution}
\end{figure}

\subsection{Effects of Token Position on Attention Sinks}
\label{sec:effects_token_level}

% \noindent \textbf{Sink position differs across the evaluated models.}
% %
% We use the teaser cases in Fig.~\ref{fig:showcase}(b) as qualitative anchors, and test whether their patterns persist in aggregate using Tab.~\ref{tab:first_token_ratio} and Fig.~\ref{fig:attention_mass_distribution}.
% %
% The Infinity example highlights a first/global-token sink; quantitatively, its position-0 token is the dominant sink in 37.52\% of evaluated units and appears in every top-5 set.
% %
% The FLUX.1 example instead places high attention near structural text--image or layout locations, not at the sequence start.
% %
% This matches the aggregate result: FLUX.1 and Z-Image rarely select position~0 ($\rho_1=0.32\%$ and $0.47\%$; $\rho_5<1\%$), while their excess attention concentrates at text positions, modality boundaries, and image-layout anchors such as FLUX pos~511/4544 and Z-Image pos~4096/4109.
% %
% Thus, the teaser is not an isolated visualization: the evaluated causal AR model follows a raw first/global-token pattern, whereas the evaluated diffusion models favor structural boundary and layout positions.
% %
% The intervention cases in Fig.~\ref{fig:showcase}(d) and Tab.~\ref{tab:sink_token_intervention} are consistent with this distinction: early removal of Infinity's first/global token is highly destructive, while removing the selected FLUX sinks has much milder final-score effects. Because Infinity's token is also an architectural global-scale token, this comparison supports sensitivity differences but does not isolate sink causality.

\noindent \textbf{Attention sinks are positionally model-dependent}.
Fig.~\ref{fig:showcase}(b) shows that Infinity-2B concentrates attention on its first and global token, whereas FLUX.1 places strong attention near structural text-image or layout positions.
The aggregate statistics in Tab.~\ref{tab:first_token_ratio} and Fig.~\ref{fig:attention_mass_distribution} show the same pattern beyond the visual examples.
The first position is rarely selected as a sink in the evaluated diffusion models with $\rho_1(0)\!<\!1\%$, while Infinity-2B shows a pronounced position-0 concentration with $\rho_1(0)\!=\!37.52\%$.
Moreover, bidirectional diffusion models favor the text region and modality boundary, as the highlighted in the Fig.~\ref{fig:attention_mass_distribution}.
This shows that T2I attention sinks do not follow a universal first-token pattern inherited from language models.
Instead, their preferred positions depend on the generation paradigm and token layout.

\noindent \textbf{Spatial corner bias in image tokens}.
The boundary and sink markers in Fig.~\ref{fig:showcase}(b) motivate a closer look at image-layout positions.
Fig.~\ref{fig:attention_mass_distribution} shows prominent image-token peaks at FLUX.1 index $4544$, the bottom-left corner, and Z-Image index $0$, the top-left corner of its image stream.
These positions are not semantic but fixed layout anchors that repeatedly absorb attention.
Infinity-2B's analogous position is its sole $1\times1$ global-scale token, which represents a coarse global canvas rather than a local patch.
Together, the visual examples and aggregate curves indicate that image-side sinks are tied to corners, boundaries, or global layout tokens.
%
% We therefore treat them as structural anchors, while leaving their precise causal role to the controlled interventions in Sec.~\ref{sec:intervention_results}.

\begin{mybox}
    \looseness -1 \textbf{Takeaways:}
    1.~Sink positions are model-dependent rather than universally tied to the first token.
    2.~Frequent sinks usually correspond to structural anchors, including global tokens, modality boundaries, and image-corner patches.
\end{mybox}

\begin{figure}[!t]
    \centering
    \includegraphics[width=\linewidth]{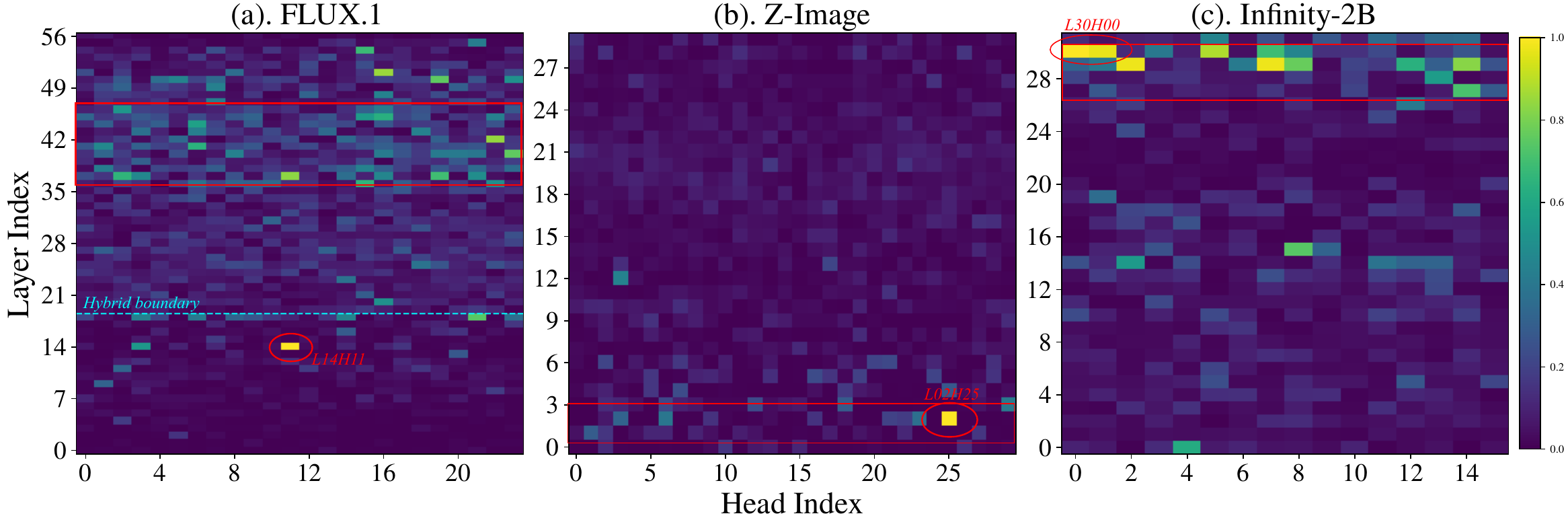}
    \caption{Layer--head maps of sink concentration $\kappa^{(l,h)}$ across models.
    Each cell represents one attention head, and higher values indicate stronger concentration on a single key token.
    Sink-dominant heads are sparse and occur at architecture-specific depths.}
    \label{fig:sink_concentration_layer_head}
\end{figure}

\subsection{Effects of Layer-Head on Attention Sinks}
\label{sec:effects_layer_head}

We use the sink concentration $\kappa^{(l,h)}$ to identify heads whose attention is dominated by a single key position.
The attention maps in Fig.~\ref{fig:showcase}(b) illustrate this unit-level view while Fig.~\ref{fig:sink_concentration_layer_head} then shows where such concentrated heads occur across the full stack.
For simplicity, we denote head $(i,j)$ as $LiHj$.

\noindent \textbf{Sink concentration is sparse across layers and heads}.
Fig.~\ref{fig:sink_concentration_layer_head} shows that only about $3\% \sim 5\%$ of heads are sink-dominant in each model.
Although their locations differ by models, this sparse localization is consistent across models.
A larger count of sink‑dominant heads in a layer increases the likelihood of sink‑token emergence, while weaker sink‑head tendencies correspondingly mitigate the sink phenomenon.
%
% Its functional relevance is explored through head pruning.
%
% The top-ranked selection causes larger average degradation than the reported random or bottom-ranked selections (Tab.~\ref{tab:sink_heads_selection}), linking the sparse maps to measurable significance.

\noindent \textbf{Sink placement occurs at model-specific depths}.
Sink depth follows each model's behavior rather than a universal pattern.
FLUX.1 peaks after the double-to-single stream transition in $L39$ to $L47$, Z-Image is dominated by the shallow head $L02H25$, and Infinity-2B concentrates sinks in middle-to-late blocks around $L27$ to $L30$.
Each model therefore exhibits a compact set of high-concentration heads at model-specific depths.
Pruning top heads changes generation scores in all three models to different degrees (Tab.~\ref{tab:intervention_head_scheme}).

\begin{mybox}
    \looseness -1 \textbf{Takeaways:}
    1.~Sink heads are sparsely localized in transformer.
    2.~Their locations differ across models rather than following a universal depth.
\end{mybox}

\subsection{Effects of Timestep on Attention Sinks}
\label{sec:effects_timestep}

% We track whether sinks become more image-directed and whether their positions persist over time.
%
% For FLUX.1 and Z-Image, $R_{\mathrm{img}}^{(p)}$ measures the share of top-5 sinks in the image region, while $R_{\mathrm{fine}}^{(p)}$ captures Infinity's shift toward finer scales.
%
% Position stability is measured by the Jaccard overlap of top-1 and top-5 sink sets across adjacent steps and full stages.

\noindent \textbf{Sink positions become increasingly image-directed}.
We analyze how sink positions evolve among the modalities during generation
Tab.~\ref{tab:sink_dynamics_timestep} shows that the image-directed ratio increases throughout generation in all three models.
FLUX.1 rises from 41.21\% to 50.47\%, Z-Image from 7.46\% to 12.73\%, and Infinity-2B from 81.24\% to 89.95\%.
This trend indicates that sink positions become more tied to visual or fine-scale structure as generation proceeds, despite differences in token layout and generation schedule.

\begin{table}[!tbp]
    \centering
    \small
    \caption{Attention sink dynamics over generation.
    \emph{Image-directed ratio} is the $R_{img}$ for FLUX.1 and Z-Image, and the $R_{fine}$ for Infinity-2B.
    \emph{Sink Position Stability} is the Jaccard overlap at adjacent steps or across each full stage.
    Larger values indicate greater image/fine-scale concentration or greater positional persistence.
    }
    \setlength{\tabcolsep}{1.3mm}
    \begin{tabular}{llccc}
        \toprule
        \multicolumn{2}{c}{Configurations}  & FLUX.1 & Z-Image & Infinity-2B \\
        \midrule
        \multicolumn{5}{c}{\textit{Image-Directed Ratio}} \\
        \midrule
        Early Stage   & $T\!\rightarrow\!0.7T$  & 41.21\% & 7.46\%  & 81.24\% \\
        Middle Stage  & $0.7T\!\rightarrow\!0.3T$ & 45.34\% & 9.70\%  & 86.61\% \\
        Late Stage    & $0.3T\!\rightarrow\!0$   & 50.47\% & 12.73\% & 89.95\% \\

        \midrule
        \multicolumn{5}{c}{\textit{Sink Position Stability}} \\
        \midrule
        \multirow{2}{*}{Adjacent Step} & Top-1  & 79.12\% & 76.03\% & 52.56\% \\
         & Top-5  & 55.10\% & 68.78\% & 29.67\% \\ \midrule
        \multirow{2}{*}{Full Stage} & Top-1      & 61.29\% & 54.11\% & 39.58\% \\
        & Top-5      & 30.23\% & 36.47\% & 22.79\% \\
        \bottomrule
    \end{tabular}
    
    \label{tab:sink_dynamics_timestep}
\end{table}

\begin{figure}[!t]
    \centering
    \includegraphics[width=\linewidth]{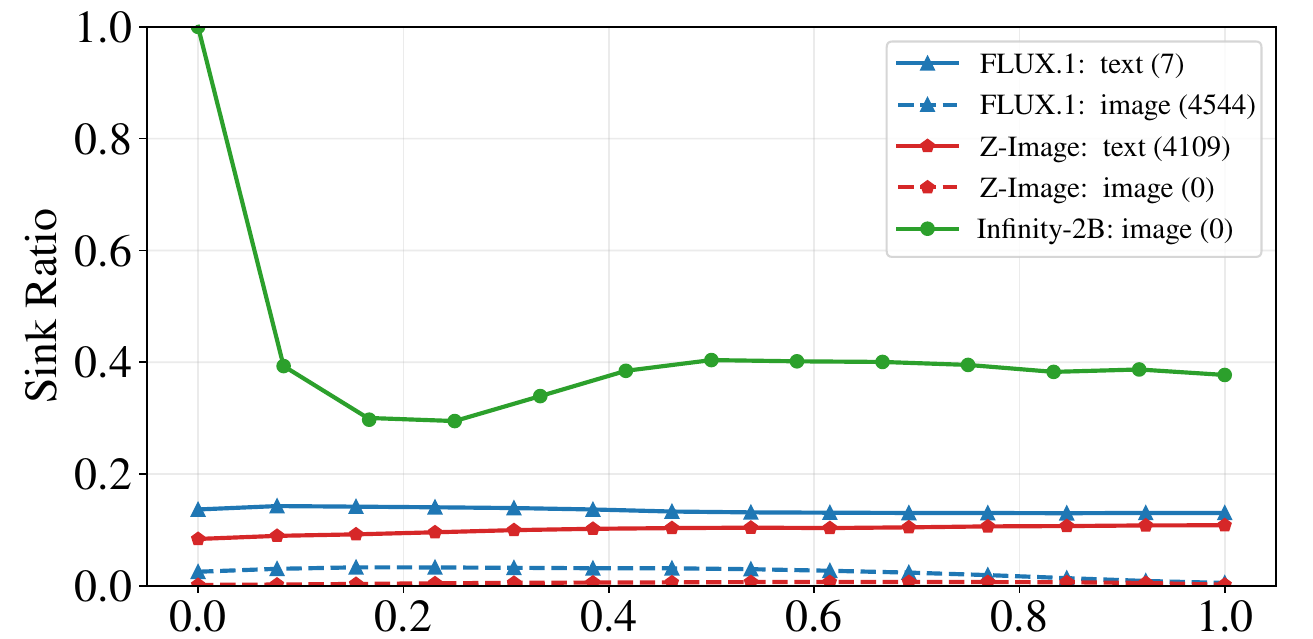}
    \caption{Trajectories of  $\rho_1^{(t)}(c)$ for text/image positions during generation.
    Candidate positions $c$ are from Fig.~\ref{fig:attention_mass_distribution}.}
    \label{fig:sink_ratio_timesteps}
\end{figure}

\noindent \textbf{Sink positions are locally stable but drift across stages}.
As shown in Tab.~\ref{tab:sink_dynamics_timestep}, sink positions remain relatively stable between adjacent steps, especially in FLUX.1 and Z-Image, where top-1 overlap reaches 79.12\% and 76.03\%.
However, the overlap decreases across full stages, which suggests gradual relocation rather than independent step-wise switching.
Moreover, Infinity-2B shows lower stability with top-1 overlap decreasing from 52.56\% to 39.58\%, consistent with its progressively expanding token set.
Overall, sinks preserve local continuity during nearby steps, but their preferred positions shift as generation moves across stages.

\begin{mybox}
    \looseness -1 \textbf{Takeaways:}
    1.~Sink positions become more image-directed as generation proceeds.
    2.~Sink positions are stable across nearby steps but drift across longer stages.
\end{mybox}

\subsection{Effects of Prompt on Attention Sinks}
\label{sec:effects_prompt}

% GenEval~\cite{geneval_ghosh2023} provides short compositional prompts, while DPG-Bench~\cite{dpg_hu2024ella} uses longer and denser prompts, allowing us to separate head stability from positional variation.

We compare GenEval~\cite{geneval_ghosh2023} and DPG-Bench~\cite{dpg_hu2024ella} to examine whether attention sinks are stable under different prompt distributions.
GenEval contains shorter compositional prompts, while DPG-Bench contains longer and denser prompts.

\noindent \textbf{Sink heads are stable across datasets}.
The layer-head concentration maps are highly correlated across the two datasets, with Pearson correlations above 0.98 for all three models, as shown in Tab.~\ref{tab:cross_data_metric}.
This indicates that the heads prone to sink behavior are largely stable under different prompt sets.
Together with the pruning results in Tab.~\ref{tab:sink_heads_selection}, these results suggest that sink-head identity is more tied to model structure than to a specific prompt benchmark.

\noindent \textbf{Sink-position entropy differs across prompt datasets}.
The normalized entropy of sink positions $H_{\mathrm{norm}}$ is computed for each model and dataset.
Although sink heads remain stable, sink positions show stronger dataset dependence.
In the diffusion models, the high-entropy ratio increases substantially on DPG-Bench, from 6.8\% to 92.2\% for FLUX.1 and from 2.7\% to 49.4\% for Z-Image.
Infinity-2B remains nearly unchanged.
This suggests that prompt distribution mainly affects where stable sink heads route attention, rather than which heads become sink-dominant.
Because the datasets differ in prompt length, semantics, and complexity, we do not attribute this change to a single prompt factor.

\begin{table}[!tb]
    \centering
    \small
    \caption{Cross-dataset stability of sink heads and sink positions.
    Pearson $r$ compares layer--head concentration maps between $\mathcal{D}_{\text{GenEval}}$ and $\mathcal{D}_{\text{DPG}}$.
    $H_{\mathrm{norm}}$ is the normalized entropy of top-1 sink positions across prompts, and $H_{\mathrm{norm}}\ge 0.5$ gives the fraction of high-entropy heads.
    Higher entropy indicates greater positional variation.}
    \setlength{\tabcolsep}{1.7mm}
    \begin{tabular}{lccccc}
    \toprule
    Model & \multicolumn{1}{l}{Sink Heads} & \multicolumn{2}{c}{$\mathrm{H}_{norm}$} & \multicolumn{2}{c}{$\mathrm{H}_{norm}\ge 0.5$} \\
     & Pearson r & $\mathcal{D}_{\text{Geneval}}$ & $\mathcal{D}_{\text{DPG}}$ & $\mathcal{D}_{\text{Geneval}}$ & $\mathcal{D}_{\text{DPG}}$ \\
     \midrule
    FLUX.1 & 0.988 & 0.37 & 0.61 & 6.8\% & 92.2\% \\
    Z-Image  & 0.983 & 0.31 & 0.43 & 2.7\% & 49.4\% \\
    Infinity-2B  & 0.999 & 0.26 & 0.26 & 2.3\% & 2.4\% \\
    \bottomrule
    \end{tabular}
    
    \label{tab:cross_data_metric}
\end{table}

\begin{mybox}
    \looseness -1 \textbf{Takeaways:}
    1.~Sink-head identities are stable across prompt distributions.
    2.~Sink positions vary more with prompt distribution, especially in diffusion models.
\end{mybox}

% !TEX root = ../main.tex
\section{Intervention Results}
\label{sec:intervention_results}

We next move from descriptive sink patterns to functional probes.
Given the sink positions and sink-dominant heads identified in Sec.~\ref{sec:systematic_analysis}, we perturb the corresponding pathways and measure generation changes relative to each model's baseline.
Because several frequent sinks also coincide with modality boundaries, corner positions, or global-scale tokens, we interpret score changes as pathway sensitivity rather than isolated causal effects of sink behavior alone.

\begin{table*}[!t]
    \centering
    \small
    \caption{Effects of three interventions on the top-50 sink heads across models.
    Logit penalizes attention to selected sink positions, Value zeros their value vectors, and Head removes the selected sink-dominant heads.
    GenEval and DPG are averages over four independent runs.
    $\bar{\Delta}$ is the mean $\%$ change from baseline over $4$ metrics, where more negative values denote greater degradation.}
    \setlength{\tabcolsep}{0.45mm}
    \begin{tabular}{c|ccccc|ccccc|ccccc}
        \toprule
        Settings~\textbackslash~Models & \multicolumn{5}{c|}{FLUX.1} & \multicolumn{5}{c|}{Z-Image} & \multicolumn{5}{c}{Infinity-2B} \\
        \midrule
        Metrics & GenEval & DPG & HPSv2 & ImgRwd & $\bar{\Delta}$(\%) & GenEval & DPG & HPSv2 & ImgRwd & $\bar{\Delta}$(\%) & GenEval & DPG & HPSv2 & ImgRwd & $\bar{\Delta}$(\%) \\
        \midrule
        Baseline & 67.14 & 83.67 & 29.62 & 0.9539 & - & 76.18 & 87.41 & 29.28 & 0.9481 & - & 64.56 & 81.84 & 29.68 & 0.9761 & - \\
        \midrule
        % \multicolumn{16}{l}{\emph{Intervention Strategy.}}\\ % ~$\|\mathcal{H}_{sink}\|=50$}
        % \midrule
        Logit. & 66.99 & 83.86 & 29.58 & 0.9478 & -0.19 & 76.55 & 87.27 & 28.92 & 0.9429 & -0.36 & 62.57 & 78.82 & 29.40 & 0.9424 & -2.79 \\
        Value. & 66.51 & 83.84 & 29.60 & 0.9343 & -0.71 & 75.98 & 87.15 & 28.93 & 0.9427 & -0.58 & 61.35 & 76.41 & 28.68 & 0.9096 & -5.45 \\
        Head. & 63.69 & 82.29 & 29.47 & 0.8636 & -4.19 & 72.94 & 87.06 & 28.98 & 0.9125 & -2.36 & 60.84 & 75.72 & 28.37 & 0.8923 & -6.56 \\
        % Head (top-100) &  &  &  &  &  &  &  &  &  &  &  &  &  &  &  \\
        % \midrule
        % \multicolumn{16}{l}{{\emph{Comparison with Random Heads.}}} \\
        % \midrule
        % top-k sink-heads & 63.69 & 82.29 & 29.47 & 0.8636 & -4.19 & 72.94 & 87.06 & 28.98 & 0.9125 & -2.36 & 60.84 & 75.72 & 28.37 & 0.8923 & -6.56 \\
        % random heads & 65.49 & 82.78 & 29.53 & 0.9325 & -1.51 & 74.02 & 87.35 & 29.13 & 0.9326 & -1.26 & 63.24 & 78.04 & 29.18 & 0.9326 & -3.19 \\
        % bottom-k sink-heads & 66.20 & 83.05 & 29.57 & 0.9367 & -1.03 & 75.34 & 87.24 & 29.08 & 0.9335 & -0.88 & 63.86 & 79.23 & 29.32 & 0.9389 & -2.32 \\
        \bottomrule
    \end{tabular}
    
    \label{tab:intervention_head_scheme}
\end{table*}

\subsection{Impact of Intervention Strategies.}
We evaluate the three strategies from Sec.~\ref{sec:intervention_strategy} using GenEval, DPG, HPSv2, and ImageReward.
Fig.~\ref{fig:showcase}(c) illustrates their intended targets, Fig.~\ref{fig:showcase}(d) shows their visual effects, and Tab.~\ref{tab:intervention_head_scheme} reports the aggregate scores.
Across models, severity follows the intervention scope.
Logit penalties produce the smallest average changes ($-0.19\%$, $-0.36\%$, and $-2.79\%$ for FLUX.1, Z-Image, and Infinity-2B), indicating limited final-score sensitivity when only routing to the selected sinks is discouraged.
Value substitution causes larger drops ($-0.71\%$, $-0.58\%$, and $-5.45\%$), consistent with the selected sink values carrying useful generation information.
Head pruning is most disruptive ($-4.19\%$, $-2.36\%$, and $-6.56\%$), as it removes both sink routing and the surrounding head computation.
Infinity-2B is consistently the most sensitive, which agrees with the first-token case in Fig.~\ref{fig:showcase}(d).

\begin{figure}[!t]
    \centering
    \includegraphics[width=\linewidth]{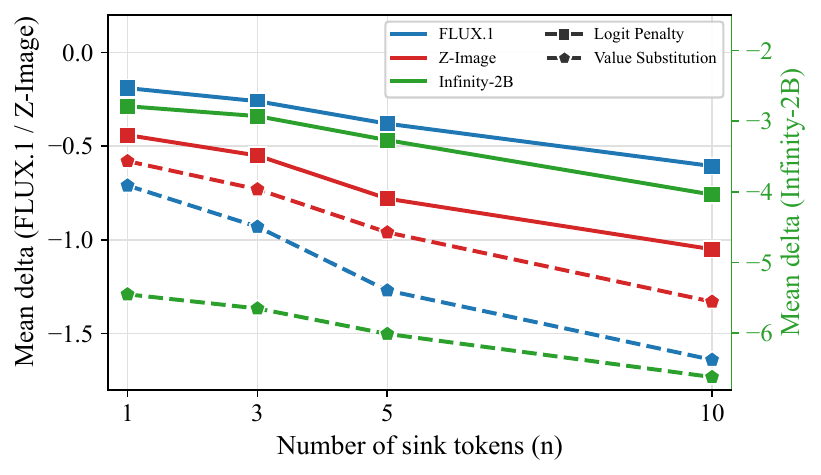}
    \caption{
        Curves of $\bar{\Delta}$ vs. number of perturbed sink tokens for logit and value intervention experiments.
        FLUX.1 and Z‑Image use left axis, Infinity‑2B right axis.
        }
    \label{fig:sink_ratio_abaltion_logit_value}
\end{figure}

\noindent \textbf{Sensitivity to sink token numbers}.
To test the effect of the perturbed sink-token count, we vary this number and plot $\bar{\Delta}$ under logit and value interventions in Fig.~\ref{fig:sink_ratio_abaltion_logit_value}.
The slope is steepest for Infinity-2B, whose dominant sink coincides with the first/global token, while FLUX.1 and Z-Image degrade more gently, matching their more distributed sinks.
This token-count ablation thus supports the positional analysis in Sec.~\ref{sec:effects_token_level}, as sink sensitivity grows with how much of the model-specific sink structure is covered.

\subsection{Sensitivity to Structural Sink-Token Removal.}
We compare structural sink-token removal across generation stages in Tab.~\ref{tab:sink_token_intervention}, with visual outputs shown in Fig.~\ref{fig:showcase}(d).
For Infinity-2B, removing position-0 token from $s1$ reduces GenEval from $64.56$ to $11.37$ and drives ImageReward from $0.9761$ to $-0.8055$; starting at $s4$ or $s8$ recovers GenEval to $62.90$ or $63.35$.
In contrast, FLUX.1 does not exhibit such degradation upon sink-token removal (positions $7$, $511$, $4544$).
Its GenEval score remains between $64.81$ and $65.45$, while DPG and ImageReward stay near or slightly above baseline.
These results indicate that the compensatory effect during generation is both model‑dependent and stage‑specific.
It fails in Infinity-2B at early steps ($s1$) but becomes effective at later steps ($s4/s8$), while FLUX.1 exhibits inherent robustness across all tested positions.

\begin{table}[!t]
    \centering
    \small
    \caption{Effects of sink-token removal over selected generation intervals.
        For FLUX.1, intervals denote denoising timestep ranges (e.g., $[T,0]$ = full trajectory).
        For Infinity-2B, intervals denote autoregressive scale ranges (e.g., $[si,s12]$ = remove the position-0 global token from scale $i$ onwards).
        $\bar{\Delta}$ is the mean percentage change across the reported metrics relative to baseline.
        }
    \setlength{\tabcolsep}{1.9mm}
    \begin{tabular}{lccccc}
    \toprule
    Settings &  Geneval & DPG & HPSv2 & ImgRwd & $\bar{\Delta}(\%)$\\
    \midrule
    FLUX.1 &  67.14 & 83.67 & 29.62 & 0.9539 & - \\
    $[T,0]$ & 64.81 & 84.19 & 29.46 & 0.9638 & -0.59 \\
    $[T,T//2]$ & 65.45 & 83.74 & 29.59 & 0.9631 & -0.39 \\
    $[T//2,0]$ & 64.95 & 83.77 & 29.48 & 0.9617 & -0.70 \\
    \midrule
    Infinity-2B & 64.56 & 81.84 & 29.68 & 0.9761 & - \\
    $[s1,s12]$ & 11.37 & 20.31 & 22.56 & -0.8055 & -91.02 \\
    $[s2,s12]$ & 22.78 & 35.73 & 24.67 & -0.3545 & -68.56 \\
    $[s3,s12]$ & 53.82 & 65.45 & 28.38 & 0.4553 & -23.60 \\
    $[s4,s12]$ & 62.90 & 73.87 & 29.20 & 0.8548 & -6.59 \\
    $[s8,s12]$ & 63.35 & 78.23 & 29.58 & 0.9727 & -1.74 \\
    \bottomrule
    \end{tabular}

    \label{tab:sink_token_intervention}
\end{table}

\subsection{Interventions on Sink Heads.}

We rank heads by their average sink concentration across timesteps and prompts to test the sparse-head hypothesis in Sec.~\ref{sec:effects_layer_head}.
If high-$\kappa$ heads are merely visual artifacts, pruning them should not be more harmful than pruning low or random heads.
On Z-Image, pruning more high-concentration heads progressively increases degradation, from Top-$20$ ($-1.12\%$) to Top-$100$ ($-5.84\%$) in Tab.~\ref{tab:sink_heads_selection}.
At a fixed budget of $50$ heads, Top-$50$ pruning causes greater average and GenEval degradation than Random-$50$ or Bottom-$50$.
This experiment confirms that the selected sink heads indeed play a more significant role than other heads during generation, providing either semantic or structural information.

\begin{table}[!thbp]
    \centering
    \small
    \caption{Ablation results of pruning selected attention heads on Z-Image.
        Top-$K$/Bottom-$50$ removes the $K$ heads with highest/lowest sink concentration, while Random-$50^*$ randomly selects $50$ heads for three runs.
        $\bar{\Delta}$ is the mean percentage change across metrics relative to baseline.
        }
    \setlength{\tabcolsep}{1.75mm}
    \begin{tabular}{llcccc}
    \toprule
    Settings &  Geneval & DPG & HPSv2 & ImgRwd & $\bar{\Delta}(\%)$\\
    \midrule
    Baseline &  76.18 & 87.41 & 29.28 & 0.9481 & - \\
    Top-20 & 75.42 & 87.21 & 29.24 & 0.9187 & -1.12 \\
    Top-50 & 72.94 & 87.06 & 28.98 & 0.9125 & -2.36 \\
    Top-100 & 71.44 & 86.87 & 28.69 & 0.8104 & -5.84 \\
    Random-50$^*$ & 74.02 & 86.91 & 29.13 & 0.9156 & -1.84 \\
    Bottom-50 & 74.32 & 87.02 & 29.07 & 0.9214 & -1.60 \\
    % Random-100 &  &  &  &  &  \\
    % Bottom-100 &  &  &  &  &  \\
    \bottomrule
    \end{tabular}
    
    \label{tab:sink_heads_selection}
\end{table}

\begin{table}[!thbp]
    \centering
    \small
    \caption{Register-token probe on FLUX.1 generation quality and position-wise top-1 sink ratios.
    Adding the register improves GenEval and DPG while slightly reducing the dominance of native sink positions.
    The low but nonzero $\rho_1(c_{\mathrm{reg}})$ supports partial redistribution of sink dominance rather than wholesale replacement of the original structural sinks.
    }
    \setlength{\tabcolsep}{0.8mm}
    \begin{tabular}{lccccc}
    \toprule
    Settings & GenEval & DPG & $\rho_1(7)$ & $\rho_1(4544)$ & $\rho_1(c_{\mathrm{reg}})$ \\
    \midrule
    FLUX.1 & 67.14 & 83.67 & 12.95\% & 3.04\% & -- \\
    + register token & 68.85 & 84.01 & 12.44\% & 2.72\% & 0.53\% \\
    \bottomrule
    \end{tabular}
    
    \label{tab:register_token_results}
\end{table}

\subsection{Probing Sink Attention with Register Tokens.}
\label{sec:register_token_experiment}

The interventions above weaken native sink tokens to test generation sensitivity.
We further ask the complementary question of whether sink attention can shift to a dedicated non-semantic position without hurting generation quality.
Following \citet{zhang2026one}, we insert one learnable register token to the FLUX.1 image-token sequence, expose it to joint attention, freeze the pretrained backbone, and remove the register output before unpatchification..
The register thus serves as an auxiliary routing slot rather than a direct image-content token.
Implementation details are in the Appendix.

Tab.~\ref{tab:register_token_results} shows that adding the register improves GenEval from $67.14$ to $68.85$ and DPG from $83.67$ to $84.01$. 
Meanwhile, the two recurring native sink positions are weakened but not removed, with $\rho_1(7)$ decreasing from 12.95\% to 12.44\%, and $\rho_1(4544)$ from 3.04\% to 2.72\%. 
These results show partial redistribution, not full replacement, of native sinks, suggesting that the added low-semantic slot redirects a small share of sink-dominant routing while keeping the original structural anchors. 
Since $\rho_1$ measures top-1 sink frequency, not total attention mass, we view this as a redistribution of sink dominance, not a full absorption of displaced attention.

% !TEX root = ../main.tex

\section{Related Work}
\label{sec:related_work}

\subsection{Text-to-Image Transformer}
Transformer-based T2I generation has developed along two complementary paradigms: \textit{diffusion transformers} and \textit{autoregressive models}.
Both of them rely heavily on multi-head attention as the core computational primitive.
The seminal Diffusion Transformer (DiT)~\cite{dit_AdaLN_peebles2023scalable} first replaced the U-Net with a pure transformer architecture, building upon foundational diffusion frameworks~\cite{ddpm_ho2020denoising,ddim_song2020,diffusion_beat_gan_dhariwal2021}.
SD3~\cite{SD3_esser2024scaling} introduced the MMDiT, which employed joint self-attention for text and image modalities.
This paradigm has since been advanced by open-source models such as FLUX.1~\cite{flux_2024}, Z-Image~\cite{cai2025zimage}, Qwen-Image~\cite{qwenimage_wu2025technicalreport,zhao2026qwenimage2}, and Hunyuan-Image~\cite{cao2025hunyuanimage}, as well as commercial systems ~\cite{imagen3_baldridge2024,seedance1_gao2025,seedream2_gong2025}.
In contrast, autoregressive Transformers treat images as discrete token sequences and generate them left-to-right.
Early works including DALL-E~\cite{ramesh2021dalle}, CogView~\cite{ding2021cogview}, and Parti~\cite{yu2022parti} established the paradigm with large-scale transformer models.
More recent efforts such as LlamaGen~\cite{sun2024llamagen} revisited next-token prediction without visual inductive biases, VAR~\cite{tian2024var} introduced next-scale prediction for faster inference, and Infinity series~\cite{han2025infinity,liu2026infinitystar} applied bit-wise quantization for efficient high-resolution generation.
Despite these architectural differences, both paradigms exhibit structured attention behaviors, motivating our analysis.

\subsection{Attention Sinks in Transformers}
Attention sinks (AS) are tokens that absorb extremely high attention.
This phenomenon arises from the sum-to-one constraint of Softmax: when no semantically relevant key exists, attention mass is forced onto specific uninformative tokens as a numerical no-op buffer~\cite{bondarenko2023quantizable,sun2024massive}.
AS were first characterized in autoregressive LLMs, where initial tokens dominate under causal masking~\cite{xiao2024efficient}.
Subsequent work empirically traces their emergence~\cite{gu2025empirialview,guo2024attention, survey_su2026attention}, exploits them for context anchoring~\cite{liu2026sinktrack}, attention redistribution~\cite{wang2025position,yu2024unveiling}, and connects them to activation outliers~\cite{sun2024massive}.
In \emph{Vision Transformers}, \citet{bondarenko2023quantizable} show that sinks are localized on low‑information background patches, not on positional tokens.
\citet{darcet2024vision} and \citet{lappe2026register} demonstrate that injecting learnable register tokens absorbs this excess attention.
In \emph{Multimodal LLMs}, visual sinks on non-semantic tokens degrade grounding~\cite{kang2025see,zhang2025shallow}, and further studies examine visual information pathways~\cite{luo2025sink} and audio-visual speech recognition~\cite{cappellazzo2025mitigating}.
\emph{Diffusion LMs} exhibit distinct features compared to autoregressive counterparts.
\citet{rulli2026attention} observe moving sinks that shift throughout the denoising process.
\citet{zhang2026one} further identify moving sinks as protecting against over‑mixing in diffusion while dynamic position hurts.
In \emph{Diffusion Models}, attention sinks have high norms \cite{jamal2026diffusion,wu2026taming}, while \citet{wu2026causalstudy} argue that they are not functional structures essential for semantic alignment.
\citet{li2026text} show that text template tokens act as implicit semantic registers.
However, the extent to which findings transfer across diffusion and autoregressive T2I Transformers remains unclear.
A unified comparison across layer, head, token-position, and generation-stage dimensions is still limited for recent text-to-image architectures.
% !TEX root = ../main.tex
\section{Conclusions}
\label{sec:conclusions}

We presented a systematic study of attention sinks in text-to-image Transformers.
Our results show that the evaluated T2I models do not all reproduce the canonical language-model first-token pattern, varying with multimodal token layouts and generation stages.
The evaluated autoregressive model exhibits strong first-token attraction, whereas the evaluated bidirectional models concentrate sink mass near modality boundaries and spatial corner positions.
Intervention experiments further show that model performance is sensitive to selected sink-related heads and positions, which motivates further study of dedicated non-semantic positions.
Overall, attention sinks provide a useful lens for interpreting and potentially controlling generative attention in T2I models.

\noindent \textbf{Limitations and Future Work}.
Our study covers three text-to-image Transformers, leaving broader architectures, scales, and training recipes for future validation. 
We only focus on inference-time behavior, while understanding how attention sinks emerge during training remains an important next step. 

{\small
\bibliographystyle{ieeenat_fullname}
\bibliography{references}
}

\end{document}